\documentclass[11pt,a4paper]{article}
\usepackage[hyperref]{emnlp2020}
\usepackage{times}
\usepackage{latexsym}

\usepackage{graphicx}
\usepackage{comment}

\usepackage{microtype}

\aclfinalcopy 

\title{Cross-Lingual Relation Extraction with Transformers}

\author{Jian Ni \and Taesun Moon \and Parul Awasthy \and Radu Florian\\
    IBM Research AI\\
    1101 Kitchawan Road, Yorktown Heights, NY 10598, USA\\
  {\tt \{nij, tsmoon, awasthyp, raduf\}@us.ibm.com}}

\date{}

\begin{document}
\maketitle

\begin{abstract}

Relation extraction (RE) is one of the most important tasks in information extraction, as it provides essential information for many NLP applications. In this paper, we propose a cross-lingual RE approach that does not require any human annotation in a target language or any cross-lingual resources. Building upon unsupervised cross-lingual representation learning frameworks, we develop several deep Transformer based RE models with a novel encoding scheme that can effectively encode both entity location and entity type information. Our RE models, when trained with English data, outperform several deep neural network based English RE models. More importantly, our models can be applied to perform zero-shot cross-lingual RE, achieving the state-of-the-art cross-lingual RE performance on two datasets ($68$-$89\%$ of the accuracy of the supervised target-language RE model). The high cross-lingual transfer efficiency without requiring additional training data or cross-lingual resources shows that our RE models are especially useful for low-resource languages.

\end{abstract}

\section{Introduction}
\label{introduction}

We live in an age of information. There are more data in electronic form than ever before, from news and articles on the Web, to electronic transactional and medical records. It is very challenging to discover and connect useful information and knowledge that is hidden in the huge amounts of data existing today. As an example, in 2020 alone, there were at least 200,000 scientific articles published on COVID-19 \cite{covid2020}. \emph{Information extraction} (IE) tackles this challenge via the study of automatic extraction of structured information from unstructured or semi-structured electronic documents. The structured information can be used to build knowledge graphs and relational databases which help us to better understand the big data.

\emph{Relation extraction} (RE) is a fundamental IE task that seeks to detect and classify semantic relationships between entities or events from text \cite{Doddington2004}. It provides essential information for many NLP applications such as knowledge base population \cite{Ji2011}, question answering \cite{Xu2016} and text mining \cite{Aggarwal2012}. For example, the entity \emph{New York City} and the entity \emph{United States} have a \emph{Part-Whole} relationship, and extraction of such relationships can help answer questions like ``What is the most populous city in the United States?"

RE models have evolved from feature-based statistical models (e.g., \citet{Kambhatla2004,Zhou2005,Li2014}), to neural network models that use pre-trained word embeddings (e.g., \citet{Zeng2014,Santos2015,Xu2015,Miwa2016,Nguyen2016,Wu2019,Soares2019}). The vast majority of RE research focuses on one language, building and tuning models with data from that language. In a globalization era, information and data are available in many different languages, so it is important to develop RE models that can operate across the language barriers.

Since annotating RE data by human for every language is expensive and time-consuming, it motivates the study of weakly supervised cross-lingual RE approaches that do not require manually annotated data for a new target language (e.g.,  \citet{Kim2012,Faruqui2015,Zou2018,Ni2019,Subburathinam2019}). The existing cross-lingual RE approaches require certain cross-lingual resources  between a source language (the language that one has annotated RE data for, usually English) and a target language or language-specific resources, such as aligned parallel corpora~\cite{Kim2012}, machine translation systems~\cite{Faruqui2015,Zou2018}, aligned word pairs~\cite{Ni2019}, or universal dependency parsers~\cite{Subburathinam2019}. Such resources may not be readily available in practice, which greatly limits the scalability of those approaches when applied to a large number of target languages.

In this paper, we propose a cross-lingual RE approach based on unsupervised pre-trained multilingual language representation models \cite{Devlin2019, Conneau2020}. The main contributions of the paper include:

\begin{itemize}

\item We develop several deep Transformer based RE models with a novel encoding scheme that can effectively encode both entity location and entity type information in the input sequence. Our English RE models outperform several deep neural network based English RE models, without using any language-specific resources such as dependency parsers or part-of-speech taggers.

\item Building on pre-trained multilingual embeddings, our English RE models can be applied to perform zero-shot cross-lingual RE for a target language without using any human annotation in the target language or any cross-lingual resources. Our models achieve the state-of-the-art cross-lingual RE performance on two datasets ($68$-$89\%$ of the accuracy of the supervised target-language RE model). The high cross-lingual transfer efficiency shows that our RE models are especially useful for low-resource languages.

\item Our RE models can be trained with data from multiple languages at the same time. The joint multilingual model performs better than monolingual models built on the same architecture but with one language at a time. In addition to higher accuracy, such a joint model also has many advantages in a production environment: simplified deployment and maintenance, the same memory/CPU/GPU footprint, and easy scalability.

\end{itemize}

We organize the paper as follows. In Section \ref{sec:task} we introduce the RE task and the framework. In Section \ref{sec:architectures} we present several deep Transformer based RE models with a novel encoding scheme. In Section \ref{sec:experiments} we evaluate the performance of the proposed RE models and compare them with several deep neural network based RE models on two datasets. We discuss related work in Section \ref{sec:relatedwork} and conclude the paper in Section \ref{sec:conclusion}.

\section{Task and Framework}
\label{sec:task}

\subsection{Relation Extraction Task}

For all pairs of entities in a sentence (or a sequence of words), the RE task is to determine whether these pairs of entities have a relationship, and if yes, classify the relationship into one of the pre-defined relation types \cite{Doddington2004}. 

Formally, let $\mathbf{s}=(w_1, w_2, ..., w_n)$ be a sentence with $n$ tokens. Let $\mathbf{e_1}=(w_{i_1},...,w_{j_1})$ and $\mathbf{e_2}=(w_{i_2},...,w_{j_2})$ be a pair of entities in the sentence, where $1\leq i_1 \leq j_1 < i_2 \leq j_2 \leq n$, with entity type $T_1$ and $T_2$, respectively. 

Suppose we have $K$ relation types. For all pairs of entities  $\mathbf{e_1}$ and $\mathbf{e_2}$, an RE model maps ($\mathbf{s}, \mathbf{e_1}, \mathbf{e_2}$, $T_1$, $T_2$) to a relation type $c \in \{0,1,2,...,K\}$ where we use type $0$ to indicate that the two entities under consideration do not have a relationship belonging to one of the $K$ relation types.

\begin{figure*}
\centering
\includegraphics[scale=0.4]{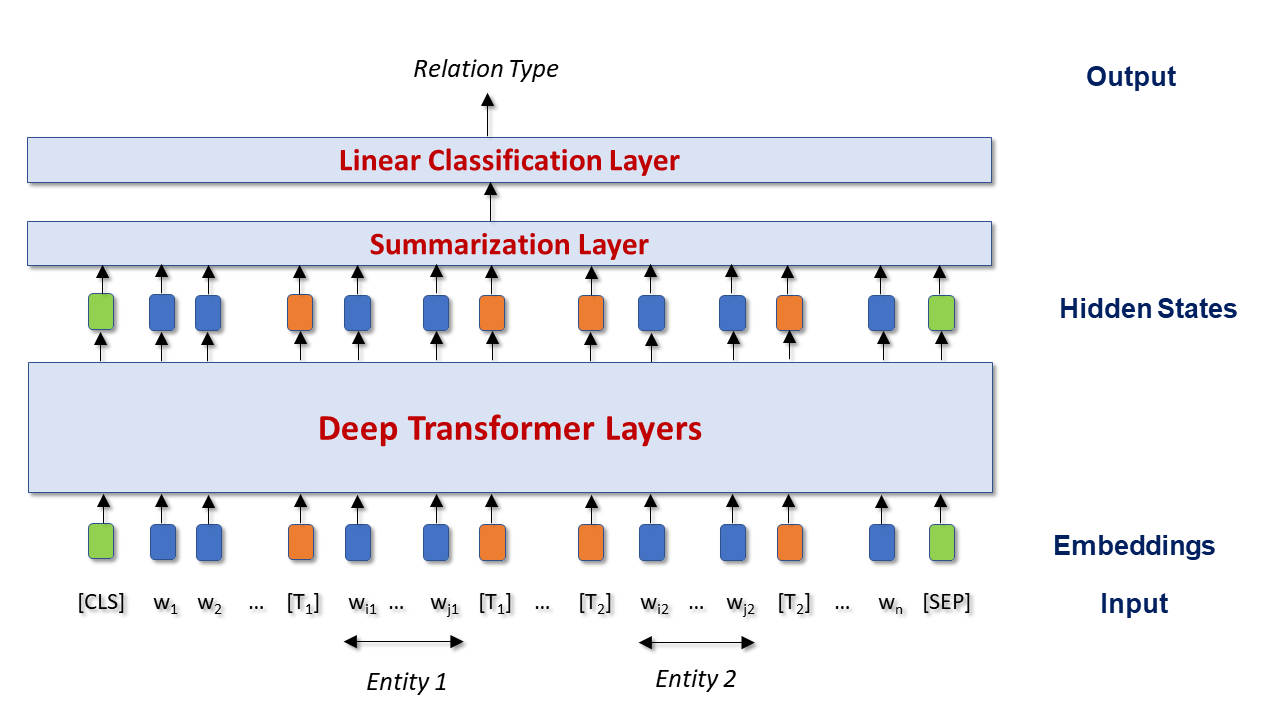} 
\caption{Deep Transformer based neural network architecture for relation extraction.} \label{fig:transformer-re-model}
\end{figure*}

\subsection{Transformer Based Language Representation Models}

The Transformer~\cite{Vaswani2017} is a neural network encoder-decoder architecture that uses a multi-head self-attention mechanism to model global dependencies between tokens in input and output sequences regardless of their distances. Since the Transformer uses only attention and avoids recurrence, it can be used to train with much longer sequences and larger datasets via significantly more parallelization compared with recurrent neural networks. 

Recent studies show that deep Transformer based language representation models, when pre-trained on a large text corpus, can achieve better generalization performance and attain the state-of-the-art performance for many NLP tasks \cite{Devlin2019,Yang2019,Liu2019,Conneau2020}. There are two phases in the framework of those models: pre-training and fine-tuning. During the \emph{pre-training} phase, a language representation model is trained on large amounts of unlabeled text. The pre-trained model parameters/embeddings are then used to initialize models for different downstream tasks, allowing effective transfer learning. During the \emph{fine-tuning} phase, the parameters of the downstream task model are fine-tuned using labeled data from the downstream task.

BERT~\cite{Devlin2019} is a language representation model based on a multi-layer bidirectional  Transformer encoder architecture. It uses a masked language model objective and a next sentence prediction objective during pre-training. RoBERTa \cite{Liu2019} improves the pre-training of BERT by training the model longer with bigger batches and more data.

Our work uses the multilingual version of BERT, named mBERT\footnote{https://github.com/google-research/bert/blob/master/multilingual.md}, and the multilingual version of RoBERTa, named XLM-R \cite{Conneau2020}.  mBERT was pre-trained with Wikipedia text of 104 languages with the largest sizes, and XLM-R were pre-trained with Wikipedia text and CommonCrawl Corpus of 100 languages. Both models use no cross-lingual resources and belong to the unsupervised representation learning framework.

\section{Transformer Based RE Models} \label{sec:architectures}

The RE task relies heavily on both syntactic and semantic information, with possibly multiple entities/relations existing in one sentence, and therefore cannot be simply formulated as a sentence classification problem. There are several issues that are special to the RE task:  

\begin{itemize}
    \item[1)] How to encode entity location information (i.e., the positions of the two entities in the sentence) into the classification model? 
    
    \item[2)] How to encode entity type information (i.e., whether an entity is of type \emph{Person}, \emph{Organization}, \emph{GPE}, etc.) into the classification model? The entity type information is important for relationship classification. For example, if we know that the two entities are of type \emph{Person}, then we can infer that they are likely to have a \emph{Personal-Social} relationship rather than an \emph{ORG-Affiliation} relationship. 
    
    \item[3)] How to create a good summary representation of the sentence and the two entities from the hidden states for the RE task?
\end{itemize}

Our deep Transformer based RE model architecture is shown in Figure \ref{fig:transformer-re-model}. We first apply an effective encoding scheme to encode both entity location and entity type information in the input sequence (Section \ref{sec:encoding}). The input tokens are then mapped to subword embeddings which are passed to deep Transformer layers (Section \ref{sec:deepTlayers}). We create a fixed-length summary representation from the hidden states of the final Transformer layer (Section \ref{sec:summarization}) and pass it to a linear classification layer (Section \ref{sec:classification}).

\subsection{Encoding of Entity Location and Type} \label{sec:encoding}

For an input sequence $\mathbf{s}=(w_1, w_2, ..., w_n)$ and two entities $\mathbf{e_1}=(w_{i_1},...,w_{j_1})$,  $\mathbf{e_2}=(w_{i_2},...,w_{j_2})$ with entity types $T_1$ and $T_2$, we first add a special token [CLS] to mark the start of the sentence and a special token [SEP] to mark the end of the sentence as in BERT \cite{Devlin2019}.

In \citet{Wu2019} and \citet{Soares2019}, special tokens are used to mark the start and end of the entities to encode entity location information:
\begin{eqnarray} \label{eq:uniform-marker-input}
    \mathbf{s}' & = & \Big([CLS], w_1, ..., [E_1], w_{i_1},...,w_{j_1}, [/E_1], ...,  \nonumber \\ 
       & & [E_2], w_{i_2},...,w_{j_2}, [/E_2], ..., w_n, [SEP]\Big)
\end{eqnarray}

In this paper, we use the entity type of an entity to mark the start and end of that entity. So we have the following modified input tokens:
\begin{eqnarray} \label{eq:entity-type-marker-input}
    \mathbf{s}'' & = & \Big([CLS], w_1, ..., [T_1], w_{i_1},...,w_{j_1}, [T_1], ...,  \nonumber \\ 
       & & [T_2], w_{i_2},...,w_{j_2}, [T_2], ..., w_n, [SEP]\Big)
\end{eqnarray}

\noindent For example, if the input tokens are: \\
\emph{``New York City is the most populous city in the United States ."}

\noindent after adding the special tokens ([CLS], [SEP]) and the entity type markers, we have the following modified input tokens: \\
\emph{``[CLS] [GPE] New York City [GPE] is the most populous city in the [GPE] United States [GPE] . [SEP]"}

The entity markers in (\ref{eq:uniform-marker-input}) are the same for entities with different types, so we call them \emph{uniform markers} (UM). In our formulation (\ref{eq:entity-type-marker-input}), the \emph{entity type markers} (ETM) encode both entity location and entity type information, killing two birds with one stone. Experiment results show that models based on entity type markers achieve higher accuracy than models based on uniform markers.

\subsection{Deep Transformer Layers} \label{sec:deepTlayers}

The modified input tokens (\ref{eq:entity-type-marker-input}) are mapped to subword embeddings (WordPiece tokenizer \cite{Wu2016} with 110k vocabulary size for mBERT and SentencePiece tokenizer \cite{Kudo2018} with 250k vocabulary size for XLM-R). The embeddings are then passed to multiple Transformer layers (12 layers for mBERT and 24 layers for XLM-R).

Each Transformer layer has two sub-layers: the first is a multi-head self-attention layer and the second is a position-wise fully connected feed-forward layer \cite{Vaswani2017}. Both sub-layers have a residual connection \cite{He2016} followed by layer normalization \cite{Ba2016}. Let the output (hidden states) of the final Transformer layer be:
\begin{eqnarray} \label{eq:entity-marker-output}
    \mathbf{H}  & = & \Big(\mathbf{h}_{[CLS]}, \mathbf{h}_1, ..., \mathbf{h}^l_{[T_1]}, \mathbf{h}_{i_1},...,\mathbf{h}_{j_1},  \mathbf{h}^r_{[T_1]}, ...,  \nonumber \\ 
     & & \mathbf{h}^l_{[T_2]}, \mathbf{h}_{i_2} ,..., \mathbf{h}_{j_2}, \mathbf{h}^r_{[T_2]}, ..., \mathbf{h}_n, \mathbf{h}_{[SEP]}\Big)  
\end{eqnarray}

\subsection{Summarization Layer} \label{sec:summarization}

In this subsection we present various schemes to create a fixed-length \emph{summary representation} of the sentence and the two entities from the hidden states (\ref{eq:entity-marker-output}) for relationship classification.

\subsubsection{Sentence Start (SS)}

The final hidden state vector for the \emph{sentence start} token [CLS], $\mathbf{h}_{[CLS]}$, contains information of all the tokens in the sentence via the attention mechanism. In this scheme, $\mathbf{h}_{[CLS]}$ is used as a summary representation of the whole sentence as in BERT \cite{Devlin2019}: 
\begin{equation}
\mathbf{h_s} = \mathbf{h}_{[CLS]}    
\end{equation}

While $\mathbf{h}_{[CLS]}$ does not explicitly contain entity location or type information, such information can be encoded in the input tokens as in (\ref{eq:entity-type-marker-input}).

\subsubsection{Entity Start (ES)}

To better incorporate the entity information into the classification model, this scheme uses the concatenation of $\mathbf{h}^l_{[T_1]}$ and $\mathbf{h}^l_{[T_2]}$, the final hidden state vectors for the two \emph{entity start} (ES) tokens, as a summary representation of the two entities: \begin{equation}
    \mathbf{h_s} = \Big[\mathbf{h}^l_{[T_1]}, \mathbf{h}^l_{[T_2]}\Big]
\end{equation}

\noindent This scheme is motivated by the entity start scheme in \cite{Soares2019} where uniform markers (\ref{eq:uniform-marker-input}) are used and the concatenation of $\mathbf{h}_{[E_1]}$ and $\mathbf{h}_{[E_2]}$ is used as a summary representation of the two entities. The difference is that $\mathbf{h}^l_{[T_i]}$ incorporates both the entity location information (the start of an entity) and entity type information since we use the entity type markers (\ref{eq:entity-type-marker-input}).

\subsubsection{Entity Max Pooling (EMP)}

In this scheme, to represent an entity, we perform element-wise \emph{max pooling} among the final hidden state vectors for the entity start token, the entity tokens and the entity end token of the two entities:
\begin{eqnarray} \label{eq:max-pool-start}
\mathbf{h}_{\mathbf{e}_1}(k)  = \max_{\mathbf{h}^l_{[T_1]}, \mathbf{h}_{i_1},...,\mathbf{h}_{j_1}, \mathbf{h}^r_{[T_1]}} \mathbf{h}(k), 1\leq k\leq H \nonumber \\
\mathbf{h}_{\mathbf{e}_2}(k)  = \max_{\mathbf{h}^l_{[T_2]}, \mathbf{h}_{i_2},...,\mathbf{h}_{j_2}, \mathbf{h}^r_{[T_2]}} \mathbf{h}(k), 1\leq k\leq H \nonumber \\
\end{eqnarray}
where $H$ is the dimension of the hidden state vectors. This scheme then concatenates the two entity representation vectors $\mathbf{h}_{\mathbf{e}_1}$ and $\mathbf{h}_{\mathbf{e}_2}$ as a summary representation of the two entities: \begin{equation}
    \mathbf{h_s} = \Big[\mathbf{h}_{\mathbf{e}_1}, \mathbf{h}_{\mathbf{e}_2} \Big]
\end{equation}

One motivation of the max-pooling operation in (\ref{eq:max-pool-start}) is to incorporate both entity type information and entity token information. In our experiments we found that this max pooling scheme achieves higher accuracy than the mention pooling scheme in \cite{Soares2019} which performs max pooling among the entity tokens only.

\subsection{Linear Classification Layer} \label{sec:classification}

The summary representation $\mathbf{h_s}$ is passed to a final linear classification layer that returns a probability distribution over the relation types:
\begin{equation} \label{eq:linearclassification}
\mathbf{p} = \textrm{softmax} \big(\mathbf{W} \mathbf{h}_s + \mathbf{b} \big)
\end{equation}

\section{Experiments} \label{sec:experiments}

In this section, we evaluate the performance of our Transformer based RE models and compare them with the RE models in previous works \cite{Ni2019,Soares2019} on two datasets.

\subsection{Datasets}

The ACE05 dataset \cite{Walker2006} includes manually annotated RE data for 3 languages: English, Arabic and Chinese. It defines 7 entity types (\emph{Person}, \emph{Organization}, \emph{Geo-Political Entity}, \emph{Location}, \emph{Facility}, \emph{Weapon}, \emph{Vehicle}) and 6
relation types  between the entities (\emph{Agent-Artifact}, \emph{General-Affiliation}, \emph{ORG-Affiliation}, \emph{Part-Whole},
\emph{Personal-Social}, \emph{Physical}).

The KLUE dataset \cite{Han2010} includes manually annotated RE data for 6 languages: English, German, Spanish, Italian, Japanese and Portuguese.
It defines 56 entity types (e.g., \emph{Person}, \emph{Organization}, \emph{Geo-Political Entity}, \emph{Location}, \emph{Facility}, \emph{Time},
\emph{Event{\_V}iolence}, etc.) and 53 relation types between the entities (e.g., \emph{Agent-Of}, \emph{Located-At}, \emph{Part-Of},
\emph{Time-Of}, \emph{Affected-By}, etc.).  

For comparison studies, we use the same data split in \cite{Ni2019}: for each language, $80\%$ of the data is selected as the training set, $10\%$ as the development set, and the remaining $10\%$ as the test set, with sizes shown in Table \ref{table:dataset-size}.
The development sets are used for tuning the model hyperparameters and for early stopping.

\begin{table}
\small
\begin{center}
\begin{tabular}{|c|c|c|c|}
\hline \textbf{ACE05} & \textbf{Train} & \textbf{Dev} & \textbf{Test} \\
\hline English (en)  & 479 & 60 & 60 \\
\hline Arabic (ar)   & 323 & 40 & 40  \\
\hline Chinese (zh)  & 507 & 63 & 63 \\

\hline \textbf{KLUE} & \textbf{Train} & \textbf{Dev} & \textbf{Test} \\
\hline English (en)     & 1137 & 140 & 140 \\
\hline German (de)      & 280 & 35 & 35  \\
\hline Spanish (es)     & 451 & 55 & 55 \\
\hline Italian (it)    & 322 & 40 & 40 \\
\hline Japanese (ja)   & 396 & 50 & 50 \\
\hline Portuguese (pt)  & 390 & 50 & 50 \\

\hline
\end{tabular}
\end{center}
\vspace{-1mm}
\caption{Numbers of documents in the train/dev/test sets of ACE05 and KLUE.}
\label{table:dataset-size}
\vspace*{-3mm}
\end{table}

\subsection{Models to Compare}

We compare the following RE models: 
\\
(a) The Convolutional Neural Network (CNN) based RE model and the Bi-Directional Long Short-Term Memory (Bi-LSTM) based RE model  \cite{Ni2019};
\\
(b) The mBERT based RE models that use uniform markers (UM) and sentence start (SS), entity start (ES), or entity max pooling (EMP) summary representation \cite{Soares2019}\footnote{The RE models in \cite{Soares2019} are based on English BERT. In our experiments we implement those models with mBERT for cross-lingual study.};
\\
(c) The mBERT based RE model that uses uniform markers (UM) and entity start (ES) summary representation plus entity type embedding at the final classification layer;
\\
(d) The mBERT based RE models that use entity type markers (ETM) and sentence start (SS), entity start (ES), or entity max pooling (EMP) summary representation;
\\
(e) The XLM-R based RE models that use uniform markers (UM) or entity type markers (ETM), and sentence start (SS), entity start (ES), or entity max pooling (EMP) summary representation.

We use HuggingFace's pytorch implementation of Transformers (mBERT, XLM-R) \cite{Wolf2019}. mBERT has $L=12$ Transformer layers, with hidden state vector size $H=768$, number of attention heads $A=12$, and $110$M parameters. XLM-R has $L=24$, $H=1024$, $A=16$, and $550$M parameters. We learn the model parameters using Adam \cite{Kingma2015}, with a learning rate $2e$-5 for mBERT based RE models, and a learning rate $1e$-5 on ACE05 and $5e$-6 on KLUE for XLM-R based RE models. We train the RE models for 10 epochs.  It took 6 hours (ACE05) and 24 hours (KLUE) to train a XLM-R model with all the training data on a NVIDIA V100 machine.

\begin{table}
\small
\begin{center}

\begin{tabular}{|l|c|c|}
\hline \textbf{Model} & \textbf{ACE05}  &  \textbf{KLUE} \\

\hline CNN \cite{Ni2019}      & 64.3  & 66.0   \\
\hline Bi-LSTM \cite{Ni2019}  & 65.5  & 67.1   \\

\hline
\hline mBERT-UM-SS \cite{Soares2019}   & 66.1  & 73.2  \\
\hline mBERT-UM-ES \cite{Soares2019}   & 67.1 &  73.5  \\
\hline mBERT-UM-EMP \cite{Soares2019}  & 67.3 & 73.5 \\
\hline mBERT-UM-ES+Entity Type  & 68.1 & 73.7 \\
\hline mBERT-ETM-SS   & 69.5 & 74.3    \\
\hline mBERT-ETM-ES   & 69.7 &  74.7  \\
\hline mBERT-ETM-EMP  & 70.3  & 74.9  \\

\hline
\hline XLM-R-UM-SS    & 70.5 & 76.2 \\
\hline XLM-R-UM-ES    & 71.6 & 76.6 \\
\hline XLM-R-UM-EMP   & 71.3 & 76.2 \\
\hline XLM-R-ETM-SS   & 74.0 & 77.1 \\
\hline XLM-R-ETM-ES   & 74.4 & 77.5 \\
\hline XLM-R-ETM-EMP  & 73.7 & 77.6 \\
\hline

\end{tabular}

\end{center}
\vspace{-1mm}
\caption{Performance ($F_1$ score) of English RE models on the ACE05 and KLUE English test data. UM stands for ``Uniform Marker", ETM for ``Entity Type Marker", SS for ``Sentence Start", ES for ``Entity Start", and EMP for ``Entity Max Pooling".} \label{table:english-re-results}
\vspace*{-3mm}
\end{table}

\begin{table*}
\small
\begin{center}
\begin{tabular}{|l|c|c|c|c|c|c|c|c|}

\hline \textbf{Model} & \multicolumn {2}{|c|}{\textbf{ACE05}} & \multicolumn {5}{|c|}{\textbf{KLUE}} & \textbf{Average}\\
\cline{2-8}   & \textbf{ar} \ & \textbf{zh}  & \textbf{de}  & \textbf{es} & \textbf{it} & \textbf{ja} & \textbf{pt}  & \\

\hline CNN \cite{Ni2019}      & 29.9 & 42.3    & 39.5 & 48.8 & 35.7 & 29.3 & 46.3   & 38.8 \\
\hline Bi-LSTM \cite{Ni2019}  & 36.4 & 46.8    & 43.8 & 50.8 & 37.6 & 28.9 & 48.4   & 41.8 \\
\hline Ensemble \cite{Ni2019}  & 35.2 & 48.6    & 44.4 & 52.7 & 38.6 & 30.2 & 49.6   & 42.8  \\

\hline
\hline mBERT-UM-SS \cite{Soares2019}  & 38.0 & 56.2   & 46.3 & 59.0 & 39.7 & 32.8 & 52.8   &  46.4 \\
\hline mBERT-UM-ES \cite{Soares2019}  & 38.1 & 58.7    & 48.2 & 60.7 & 41.9 & 33.7 & 55.4   & 48.1 \\
\hline mBERT-UM-EMP  \cite{Soares2019}  & 38.2 & 59.7    &  47.6 & 60.9 & 40.9 & 34.6 & 54.8 & 48.1 \\ 
\hline mBERT-UM-ES+Entity Type & 38.9 & 59.0   & 48.4 & 60.8 & 41.6 & 34.0 & 55.2 & 48.3 \\

\hline mBERT-ETM-SS         & 40.0 & 59.1    & 48.5 & 61.4 & 42.0 & 36.3 & 56.3  &  49.1 \\
\hline mBERT-ETM-ES & 42.1 & 61.2    & 49.7 &  63.2 & 43.8 & 37.0 & 57.7  &  50.7 \\
\hline mBERT-ETM-EMP & 42.4 & 62.9   & 49.2 & 63.1 & 42.9 & 37.6 & 56.7 &   50.7 \\

\hline
\hline XLM-R-UM-SS  & 43.9 & 60.7 &  52.9 & 66.2 & 48.1 & 44.3 & 59.2 & 53.6 \\
\hline XLM-R-UM-ES  & 46.8 & 62.9 &  53.6 & 67.0 & 48.0 & 43.9 & 59.7 & 54.5 \\
\hline XLM-R-UM-EMP & 44.9 & 63.5 &  52.9 & 66.2 & 48.1 & 44.3 & 59.2 & 54.2 \\
\hline XLM-R-ETM-SS & 45.9 & 63.0 &  54.4 & 68.8 & 49.3 & 45.8 & 60.8 & 55.4 \\
\hline XLM-R-ETM-ES & 46.2 & \textbf{65.7} &   \textbf{55.6} & 69.4 & \textbf{50.4} & 46.5 & 61.9 & 56.5  \\ 
\hline XLM-R-ETM-EMP & \textbf{49.7} & 64.9 &   55.0 & \textbf{70.3} & \textbf{50.4} & \textbf{47.1} & \textbf{62.0} & \textbf{57.1} \\

\hline XLM-R-ETM-EMP (normalized by Supervised) & $68\%$ & $87\%$ &  $80\%$  & $89\%$ & $86\%$ & $71\%$ & $87\%$ & $81\%$ \\

\hline
\hline \emph{XLM-R-ETM-EMP (Supervised)}  & \emph{72.9} & \emph{75.0} & \emph{68.4} & \emph{78.8} & \emph{58.8} & \emph{66.1} & \emph{71.1} & \emph{70.2} \\
\hline
\end{tabular}
\end{center}
\vspace{-1mm}
\caption{Cross-lingual RE performance ($F_1$ score) of English (source-language) RE models on the ACE05 and KLUE target-language test data. UM stands for ``Uniform Marker", ETM for ``Entity Type Marker", SS for ``Sentence Start", ES for ``Entity Start", and EMP for ``Entity Max Pooling".} \label{table:cross-lingual-re-results}

\vspace*{-3mm}
\end{table*}

\subsection{English RE Model Performance}

To evaluate cross-lingual RE performance, we first use English as the source language and other languages as the target languages. 
We build supervised English RE models trained with English training data only. For each Transformer based RE model architecture, we train 5 models initiated with different random seeds, and the reported performance ($F_1$ score) in Tables \ref{table:english-re-results}, \ref{table:cross-lingual-re-results}, \ref{table:ace-per-type-f1}, \ref{table:cross-lingual-all-pairs}, \ref{table:multilingual-re-results} is averaged over the 5 models\footnote{We report the average performance score of a neural network based RE model trained with different random seeds, motivated by observations in \cite{Reimers2017}, which showed that reporting single performance scores may be insufficient to compare neural network based models.}. 

As shown in Table \ref{table:english-re-results}, deep Transformer based RE models achieve much better performance than the CNN or Bi-LSTM based RE models in \cite{Ni2019}.
Among the Transformer based RE models, here are the key observations: 
\\
(1) The models with entity type markers which encode both entity location and entity type information (e.g., mBERT-ETM-ES) outperform the models with uniform markers which encode only entity location information (e.g., mBERT-UM-ES \cite{Soares2019}).
\\    
(2) It is more effective to encode the entity type information in the input sequence and let the information be propagated to the final classification layer through the deep Transformer layers (as in mBERT-ETM-ES), than to add the entity type information directly at the final classification layer (as in mBERT-UM-ES+Entity Type).
\\
(3) The deeper XLM-R based RE models (24 Transformer layers) further improve the mBERT based RE models (12 Transformer layers).

\subsection{Cross-Lingual RE Performance}

We apply the English RE models to other target languages. For the English RE models in \cite{Ni2019}, cross-lingual model transfer is achieved by projecting the target-language word embeddings into the English embedding space via a bilingual word embedding mapping (\emph{the cross-lingual representation projection framework} \cite{Mikolov2013b,Ni2017}). For the Transformer based English RE models, since we train the models using the mBERT or XLM-R subword embeddings which were pre-trained in the same embedding space (\emph{the cross-lingual representation learning framework}), those models can be directly applied to perform zero-shot cross-lingual RE on other languages.

\begin{table}
\scriptsize
\begin{center}
\begin{tabular}{|l|c|c|c|c|c|c|}

\hline
Relation Type & \multicolumn {3}{|c|}{mBERT-UM-ES} & \multicolumn {3}{|c|}{mBERT-ETM-ES} \\ 

\cline{2-7} & en & ar & zh & en & ar & zh \\ \hline
\emph{Agent-Artifact}       & 56.3 & 32.6 & 43.4  & 59.7 & 29.2 & 47.0 \\
\emph{General-Affiliation}  & 50.0 & 13.7 & 57.5  & 51.8 & 20.4 & 59.9 \\
\emph{ORG-Affiliation}      & 75.8 & 51.6 & 71.7  & 79.6 & 56.7 & 73.9 \\
\emph{Part-Whole}           & 66.5 & 46.5 & 58.3  & 67.7 & 48.3 & 60.5 \\
\emph{Personal-Social}      & 69.3 & 25.7 & 21.2  & 71.7 & 25.3 & 22.2 \\
\emph{Physical}             & 61.1 & 23.0 & 48.2  & 62.3 & 25.4 & 51.7 \\
\hline
\end{tabular}
\end{center}
\vspace{-1mm}
\caption{Per type $F_1$ scores of the English mBERT-UM-ES and mBERT-ETM-ES models on the ACE05 development sets. See~\citet{Walker2006}
for details on the relation types.}
\label{table:ace-per-type-f1}
\vspace*{-3mm}
\end{table}

\begin{table*}
\scriptsize
\begin{center}
\begin{tabular}{|c|p{7.5cm}|c|c|c|}

\hline
\# & Text & Ground Truth & mBERT-UM-ES &   mBERT-ETM-ES \\ 
\hline
1 & \texttt{The Cleveland Cavaliers also formally introduced Paul Silas as [their]/ORGANIZATION [coach]/PERSON Monday.} & \emph{ORG-Affiliation} & \emph{Personal-Social} & \emph{ORG-Affiliation} \\

\hline
2 & \texttt{Therefore, on August 20, 2003, bravely Beatriz walked into the [USCF]/ORGANIZATION [Offices]/FACILITY in New Windsor and immediately fired 17 staff members.}  &  \emph{Agent-Artifact} &  \emph{Part-Whole}  & \emph{Agent-Artifact} \\

\hline
3 & \texttt{At [SUNY Albany]/ORGANIZATION in [NY]/GPE, about as state school as you can get, we get about 500 less than what you're talking about at Pitt. } & \emph{General-Affiliation} & \emph{Part-Whole} & \emph{General-Affiliation}  \\
\hline

\end{tabular}
\end{center}
\vspace{-1mm}
\caption{Relation extraction examples of the mBERT-UM-ES model and the mBERT-ETM-ES model on the ACE05 English development set. The two entities are in brackets with entity types following the slash.}
\label{tab:ace-examples}
\vspace*{-3mm}
\end{table*}

As shown in Table \ref{table:cross-lingual-re-results}, our deep Transformer based RE models attain much higher accuracy than the models in the previous works, achieving the state-of-the-art cross-lingual RE performance for all the languages. In particular, averaged over the 7 target languages, the XLM-R-ETM-EMP model achieves $57.1$ $F_1$ score, which is $14.3$ $F_1$ points better than the Ensemble model in \cite{Ni2019} and $9.0$ $F_1$ points better than the mBERT-UM-ES model in \cite{Soares2019}.

Among the XLM-R-ETM models, the XLM-R-ETM-SS model that uses global sentence representation does not perform as well as the XLM-R-ETM-ES or XLM-R-ETM-EMP models that use local entity representations. We tried to concatenate local entity representations (ES or EMP) with global sentence representation (SS), and we found that this does not improve the accuracy.

 For each target language, we also provide the performance of the supervised XLM-R-ETM-EMP model trained with target-language training data, which serves as an upper bound for the cross-lingual RE performance. The average cross-lingual RE performance of the English XLM-R-ETM-EMP model (57.1 $F_1$ score) reaches $81\%$ of the average performance of the supervised RE models (70.2 $F_1$ score), which is quite impressive given that some target languages (Arabic, Chinese and Japanese) are not from the same language family or using the same script as English.

\subsubsection{The Effect of Word Order}

In Table \ref{table:cross-lingual-re-results} we provide the \emph{normalized cross-lingual RE accuracy} from English to every target language under the XLM-R-ETM-EMP model (the cross-lingual performance divided by the supervised performance), to study cross-lingual RE model \emph{transfer efficiency} from English to the target languages. 

It turns out that \emph{word order} plays a key role in determining the cross-lingual transfer efficiency. English belongs to the SVO language family where in a sentence the Subject is followed by the Verb, and the Verb is followed by the Object. Chinese, Spanish, Italian and Portuguese are also SVO languages, and the normalized cross-lingual RE accuracy from English to these languages reaches nearly $90\%$, even for Chinese that uses logographic script which is totally different from English that uses Latin script. Arabic belongs to the VSO (Verb followed by Subject followed by Object) language family and Japanese belongs to the SOV (Subject followed by Object followed by Verb) language family. The normalized cross-lingual RE accuracy from English to these two languages is much lower, around $70\%$. Interestingly, German is a hybrid SVO/SOV language, with SVO in some cases and SOV in others, and the normalized cross-lingual RE accuracy from English to German is $80\%$.

For the representation projection approach in \cite{Ni2019}, it was also observed that word order affects the cross-lingual RE model transfer efficiency. The best model in \cite{Ni2019} achieves an average transfer efficiency of $70\%$ (averaged over the 7 target languages), while the XLM-R-ETM-EMP model achieves an average transfer efficiency of $81\%$ --- a substantial improvement.

\subsubsection{Uniform Markers vs. Entity Type Markers}

We analyze the differences between the mBERT-UM-ES model and the mBERT-ETM-ES model in more details. First, we provide type-level $F_1$ scores on the ACE05 development data in Table~\ref{table:ace-per-type-f1}. The mBERT-ETM-ES model has consistently better accuracy across almost all of the relation types, with 3+ $F_1$ point improvement for \emph{Agent-Artifact} and \emph{ORG-Affiliation} on the English development set. We see bigger improvements of the mBERT-ETM-ES model on some relation types for the cross-lingual RE model transfer. For example, it has 6.7 $F_1$ point improvement for \emph{General-Affiliation} and 5.1 $F_1$ point improvement for \emph{ORG-Affiliation} on the Arabic development set. This shows that entity type markers are very helpful to the Transformer based RE models, both in the monolingual scenario and in the cross-lingual scenario.

We list some examples in Table \ref{tab:ace-examples}. In the first example, the two entities in brackets (with entity types following the slash) \texttt{... [their]/ORGANIZATION [coach]/PERSON ...} have
the relation \emph{ORG-Affiliation} which was falsely labeled by mBERT-UM-ES as \emph{Personal-Social}, while mBERT-ETM-ES, with entity type information encoded, succeeded in producing the correct label.

\subsubsection{All-Pair Cross-Lingual RE Performance}

\begin{table}
\small

\begin{center}

\includegraphics[scale=0.21]{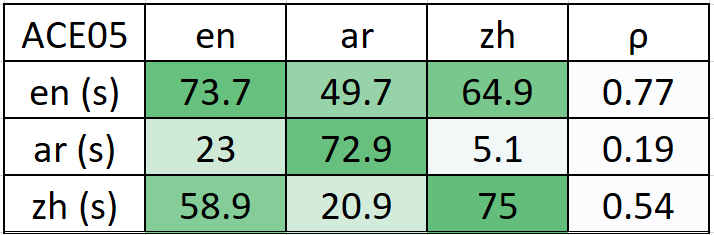} 
\includegraphics[scale=0.33]{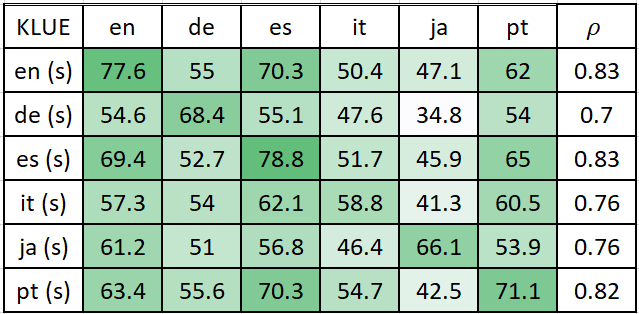} 

\end{center}
\vspace{-1mm}
\caption{All-pair cross-lingual RE performance of XLM-R-ETM-EMP: rows are source languages (s) and columns are target languages.
$\rho$ is the average normalized cross-lingual RE accuracy.}
\label{table:cross-lingual-all-pairs}
\vspace*{-3mm}
\end{table}

\begin{table*}
\small
\begin{center}
\begin{tabular}{|l|c|c|c|c|c|c|c|c|c|c|}

\hline \textbf{Model} & \multicolumn {3}{|c|}{\textbf{ACE05}} & \multicolumn {6}{|c|}{\textbf{KLUE}} & \textbf{Ave}\\
\cline{2-10}   & \textbf{en} & \textbf{ar} \ & \textbf{zh}  & \textbf{en} & \textbf{de}  & \textbf{es} & \textbf{it} & \textbf{ja} & \textbf{pt}  & \\

\hline mBERT-ETM-EMP (Monolingual) & 70.3 & 70.9 & 71.7 &   74.9 &   63.6 & 75.0 & 53.6 & 59.5 & 66.6 &  67.3  \\

\hline mBERT-ETM-EMP (Multilingual) & 70.7 & 71.7 & 73.2 &   75.2 & 65.1 & 76.1 & 55.3 & 61.3 & 68.3 &  68.5 \\

\hline XLM-R-ETM-EMP (Monolingual) &  73.7 & 72.9 & 75.0 &   77.6 & 68.4 & 78.8 & 58.8 & 66.1 & 71.1 & 71.4  \\

\hline XLM-R-ETM-EMP (Multilingual) & 73.7 &  73.5 & 75.8 &    78.3 & 68.7 & 79.7 & 60.7 & 67.4 & 71.4 & 72.1 \\

\hline
\end{tabular}
\end{center}
\vspace*{-1mm}
\caption{Comparisons of monolingual RE models (for each language, a separate model is trained with data from that language) and multilingual RE models (a joint model is trained with data from all the languages).} \label{table:multilingual-re-results}
\vspace*{-3mm}
\end{table*}

In Table \ref{table:cross-lingual-all-pairs} we provide cross-lingual RE performance from each language $s$ (as the source language) to every other language $t$ (as the target language) under the XLM-R-ETM-EMP model trained with training data of $s$.

To study the cross-lingual RE model transfer efficiency of a source language $s$ in a set of languages $\mathcal{L}$, we define $\rho_{\mathcal{L}}(s)$ to be the average normalized cross-lingual RE accuracy from $s$ to other languages in $\mathcal{L}$:
\begin{equation}
    \rho_{\mathcal{L}}(s) = \frac{\sum_{t\in \mathcal{L},t \ne s}\frac{f(s,t)}{f(t,t)}}{|\mathcal{L}|-1}
\end{equation}
where $f(s,t)$ is the cross-lingual RE performance ($F_1$ score) from source language $s$ to target language $t$, and $f(t,t)$ is the supervised RE performance of target language $t$. Larger $\rho$ indicates higher cross-lingual RE performance (transfer efficiency). As shown in Table \ref{table:cross-lingual-all-pairs}, for both datasets, English has the highest cross-lingual transfer efficiency, so it is a good choice for the source language.

\subsection{Multilingual RE Performance}

In this subsection we investigate the capability of our Transformer based RE models for handling data from multiple languages. We train a joint \emph{multilingual} RE model with data from all the languages, and compare it with \emph{monolingual} RE models where a separate model is trained with data from each language using the same architecture. Since the languages can be quite different, it is not clear whether a multilingual model can achieve better or even comparable performance as the monolingual models.

We compare the performance of the multilingual mBERT-ETM-EMP model and the multilingual XLM-R-ETM-EMP model with the corresponding monolingual models in Table \ref{table:multilingual-re-results}. 
The multilingual RE models attain consistently better accuracy than the monolingual RE models, even when the languages are from different language families and use different scripts. In addition to better accuracy, a joint model also has many advantages in a production
environment, such as simplified deployment and maintenance, the same memory/CPU/GPU footprint, and easy scalability.

\section{Related Work}
\label{sec:relatedwork}

The existing weakly supervised cross-lingual RE approaches require certain cross-lingual or language-specific resources. \citet{Kim2012} uses aligned parallel corpora to create weakly labeled RE data. \citet{Faruqui2015} and \citet{Zou2018} apply  machine translation systems to translate sentences between English and a target language. \citet{Ni2019} uses aligned word pairs to learn a bilingual word embedding mapping for representation projection. \citet{Subburathinam2019} applies universal dependency parsers to convert sentences of different languages into language-universal tree structures. Our cross-lingual RE approach does not require such resources.

A few BERT based models have been developed for the RE task. \citet{Wu2019} and \citet{Soares2019} use uniform tokens to mark the start and end of the two entities. Those models do not encode entity type information and focus on English RE.

mBERT and XLM-R have been applied to several other NLP tasks including named entity recognition (NER), part-of-speech (POS) tagging, dependency parsing and natural language inference (NLI) \cite{Pires2019,WuDredze2019,Moon2019,Conneau2020}. Unlike those tasks, the relation extraction task relies heavily on both syntactic and semantic information. We showed that mBERT/XLM-R can indeed represent such information in a language universal way and the RE models built on top of it can transcend language barriers well.

\section{Conclusion}
\label{sec:conclusion}

In this paper, we proposed a cross-lingual RE approach based on unsupervised cross-lingual representation learning frameworks.  We developed several deep Transformer based RE models with a novel encoding scheme that effectively encodes both entity location and entity type information. Our RE models can be applied to perform zero-shot cross-lingual RE, achieving the state-of-the-art cross-lingual RE performance ($68$-$89\%$ of the accuracy of the supervised target-language RE model), even in cases where the target languages are from different language families and use different scripts, without using any human annotation in the target languages or any cross-lingual/language-specific resources.

\bibliography{cross-lingual-re}
\bibliographystyle{acl_natbib}

\end{document}